\pdfoutput=1

\documentclass[11pt]{article}

\usepackage[final]{acl}

\usepackage{times}
\usepackage{latexsym}

\usepackage[T1]{fontenc}

\usepackage[utf8]{inputenc}

\usepackage{microtype}

\usepackage{inconsolata}

\usepackage{graphicx}

\title{BrainStorm @ iREL at \#SMM4H 2024: Leveraging Translation and Topical Embeddings for Annotation Detection in Tweets}

\author{
 \textbf{Manav Chaudhary\textsuperscript{1}},
 \textbf{Harshit Gupta\textsuperscript{1}},
 \textbf{Vasudeva Varma\textsuperscript{1}}
\\
 \textsuperscript{1}IIIT Hyderabad
\\
}

\begin{document}
\maketitle
\begin{abstract}
The proliferation of LLMs in various NLP tasks has sparked debates regarding their reliability, particularly in annotation tasks where biases and hallucinations may arise. In this shared task, we address the challenge of distinguishing annotations made by LLMs from those made by human domain experts in the context of COVID-19 symptom detection from tweets in Latin American Spanish. This paper presents BrainStorm @ iREL's approach to the SMM4H 2024 Shared Task, leveraging the inherent topical information in tweets, we propose a novel approach to identify and classify annotations, aiming to enhance the trustworthiness of annotated data.
\end{abstract}

\section{Introduction}

Data annotation, essential for improving machine learning models, involves labeling raw data with relevant information. However, this process is often costly and time-consuming. In recent times, the field of Natural Language Processing (NLP) has seen a transformative shift with the widespread adoption of Large Language Models (LLMs) like GPT-4 (\citet{openai2024gpt4}), Gemini (\citet{team2023gemini}) and BLOOM (\citet{le2023bloom}). These advanced models have shown remarkable capabilities in automating data annotation (\citet{tan2024large}), aiding in a crucial yet labor-intensive step in machine learning workflows. However, despite their impressive performance, the integration of LLMs in annotation tasks has sparked a debate within the research community. Proponents highlight their efficiency and consistency, while skeptics point to potential issues such as underlying biases and hallucinations.

While many recent efforts have focused on distinguishing between human and machine-generated text (\citet{hans2024spotting}, \citet{gambetti2023combat}, \citet{abburi2023simple}), detecting whether annotations are done by LLMs offers a novel perspective on AI detection. The advent of powerful LLMs, while driving innovation, poses risks of increased spread of untruthful news, fake reviews, and biased opinions, highlighting the need for a variety of detection technologies.

This paper addresses Task 7 of the SMM4H-2024 (\citet{smm4h-2024-overview}): The 9th Social Media Mining for Health Research and Applications Workshop and Shared Tasks, focusing on the identification of data annotations made by LLMs versus those made by human domain experts. Our objective is to develop methods for distinguishing between annotations made by LLMs and those by human experts in the context of COVID-19 tweets in Latin American Spanish. This task is crucial for evaluating the generalizability and reliability of LLMs in real-world applications, particularly in health-related NLP tasks.

\section{Methodology}

Our approach to identifying whether a tweet was labeled as containing COVID-19 symptoms by an LLM or a human domain expert involves several key steps. We begin by preparing the dataset and leveraging both original and translated tweet texts to evaluate the performance of different models. Additionally, we incorporate topical embeddings to enhance the distinction between human and LLM annotations.

\begin{figure*}[ht]
    \centering
    \includegraphics[width=\textwidth]{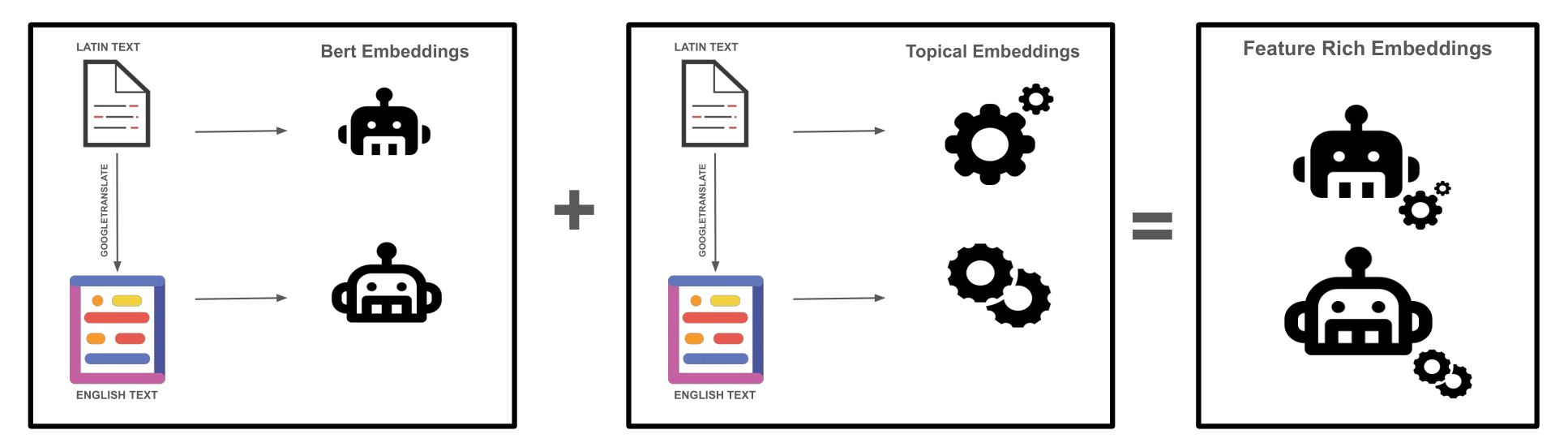}
    \caption{Diagram illustrating our method. The process starts with data translation from Latin American Spanish to English. These two datasets are used to generate BERT embeddings, followed by topical embeddings using BERTopic. These two embeddings are combined to give a new feature-rich embedding to be used for training our models.}
    \label{fig:method_diagram}
\end{figure*}

\subsection{Dataset Preparation}

The dataset consists of three columns: indexN, TweetText, and label. The TweetText column contains tweets written in Latin American Spanish, and the label column indicates whether the tweet was annotated by a human (human) or by GPT-4 (machine). The task is to determine if a tweet labeled as containing COVID-19 symptoms was annotated by an LLM or a human.

Given the bilingual nature of our approach, we first translate the Latin American Spanish tweets into English using Google Translate. This step enables us to apply and compare models trained on different languages and specific to tweet data.

\subsection{Models Used}

We compare the performance of two models:
\begin{itemize}
    \item \textbf{dccuchile/bert-base-spanish-wwm-cased (\citet{CaneteCFP2020})}: A BERT model pre-trained on Spanish text, which we use to process the original Spanish tweets. Note that we also use this model as our baseline, achieving a score of 0.50 on the test set.
    \item \textbf{vinai/bertweet-covid19-base-cased (\citet{bertweet})}: A BERT model pre-trained specifically on English COVID-19 related tweets, which we use to process the translated English tweets. Note that an ablation study using just the translated tweets and the BERTweet model have been left for future exploration. 
\end{itemize}
    
By comparing these models, we aim to leverage the strengths of language-specific and domain-specific pre-training.

\subsection{Topical Embeddings with BERTopic}

To improve the annotations further, we incorporate topical embeddings. The text data in both languages undergoes topic modeling using the BERTopic (\citet{grootendorst2022bertopic}) library. BERTopic extracts latent topics from the text using BERT embeddings. This step assigns a topic label to each tweet in both Spanish and English versions. During tokenization, the embeddings of these topic labels are appended to the tokenized representations of the tweets. Using a custom architecture, the topic embeddings are concatenated with the pooled output of the models, and the resulting combined representation is passed through a classification layer to predict the tweet's label. 

The rationale behind this is that tweets written by humans have an intrinsic topical coherence that can be captured and distinguished from machine annotations. Our hypothesis is that human-annotated tweets are more contextually consistent and thematically structured compared to those annotated by an LLM.

In our approach, we treat human annotations as the gold standard—the absolute truth. This means we assume that any tweet labeled by a human is correctly annotated. Conversely, we recognize that tweets labeled by the LLM may include both correct and incorrect annotations.

\begin{table}[h]
\centering
\caption{Classification Results on the Test Set}
\label{tab:classification_results}
\begin{tabular}{|l|l|l|l|}
\hline
Model & Score \\ \hline
Baseline Spanish & 0.50 \\
Topical Spanish & 0.50 \\ 
Topical English & 0.51 \\ \hline

\end{tabular}
\end{table}

The motivation for using Topic Modeling is based on the nature of the tweets themselves. Since the tweets are written by humans, there is an inherent topical structure that a model can learn. By utilizing topical embeddings, we enhance the model's ability to capture this structure, thus improving its performance in identifying whether the annotations were made by a human or an LLM.

\section{Results}

We evaluated the models on the test set using the accuracy score provided by the organizers on CodaLab. We observe that Topical Spanish (with BERTopic) achieved a score of 0.50, indicating that the incorporation of topical embeddings did not improve the performance over the baseline in the original Spanish tweets.

Topical English (translated tweets with BERTopic) achieved a score of 0.51, showing a marginal improvement over the baseline, suggesting some potential in the use of translated tweets and topical embeddings.

While the results indicate only slight improvements, they underscore the challenges inherent in distinguishing between human and LLM annotations in this specific context.

\section{Conclusion}

This study explored the feasibility of distinguishing between human and LLM annotations in COVID-19 symptom detection from tweets in Latin American Spanish. By leveraging both language-specific and domain-specific models, along with topical embeddings, we aimed to enhance the accuracy of annotation classification. Our findings reveal that while topical embeddings and the use of translated tweets offer some promise, the improvements are marginal. The results suggest that more sophisticated techniques or additional features might be necessary to achieve significant enhancements in performance.

The slight improvement observed with translated English tweets suggests that the method has potential when combined with domain-specific models like BERTweet, pointing to the importance of further exploring multilingual and domain-adaptive approaches. There is a need to conduct detailed ablation studies to isolate the impact of various components, such as the translation process, topical embeddings, and different pre-trained models. An investigation into more advanced topic modeling techniques or the integration of other context-aware embeddings will also help.

By addressing these areas, we can further enhance the reliability of distinguishing between human and machine annotations, ultimately contributing to more trustworthy NLP systems in critical domains like healthcare.

\bibliography{custom}

\end{document}